%% file: main.tex
\documentclass[10pt,twocolumn,letterpaper]{article}

\usepackage{cvpr}              
\input{preamble}
\definecolor{cvprblue}{rgb}{0.21,0.49,0.74}
\usepackage[pagebackref,breaklinks,colorlinks,allcolors=cvprblue]{hyperref}
\usepackage{multirow} 
\usepackage{stfloats}

\newcommand{\methodname}{WorldCanvas}
\newcommand{\method}{\texttt{\methodname}\xspace}

\def\paperID{9216} 
\def\confName{CVPR}
\def\confYear{2026}

\title{ The World is Your Canvas:\\ Painting Promptable Events with Reference Images, Trajectories, and Text}

\author{Hanlin Wang$^{1,2}$\quad Hao Ouyang$^{2}$\quad Qiuyu Wang$^{2}$\quad Yue Yu$^{1,2}$\quad Yihao Meng$^{1,2}$, \\ Wen Wang$^{3,2}$\quad Ka Leong Cheng$^{2}$\quad Shuailei Ma$^{4,2}$\quad Qingyan Bai$^{1,2}$\quad Yixuan Li$^{5,2}$, \\ Cheng Chen$^{6,2}$\quad Yanhong Zeng$^{2}$\quad Xing Zhu$^{2}$\quad Yujun Shen$^{2}$\quad Qifeng Chen$^{1\dagger}$ \\[0.2cm] $^1$HKUST \quad $^2$Ant Group \quad $^3$ZJU \quad $^4$NEU \quad $^5$CUHK \quad $^6$ NTU
}
\vspace{-3pt}
\begin{document}
\twocolumn[{
    \renewcommand\twocolumn[1][]{#1}
    \maketitle
    \begin{center}
      \vspace{-3pt}
      \includegraphics[width=0.92\textwidth]{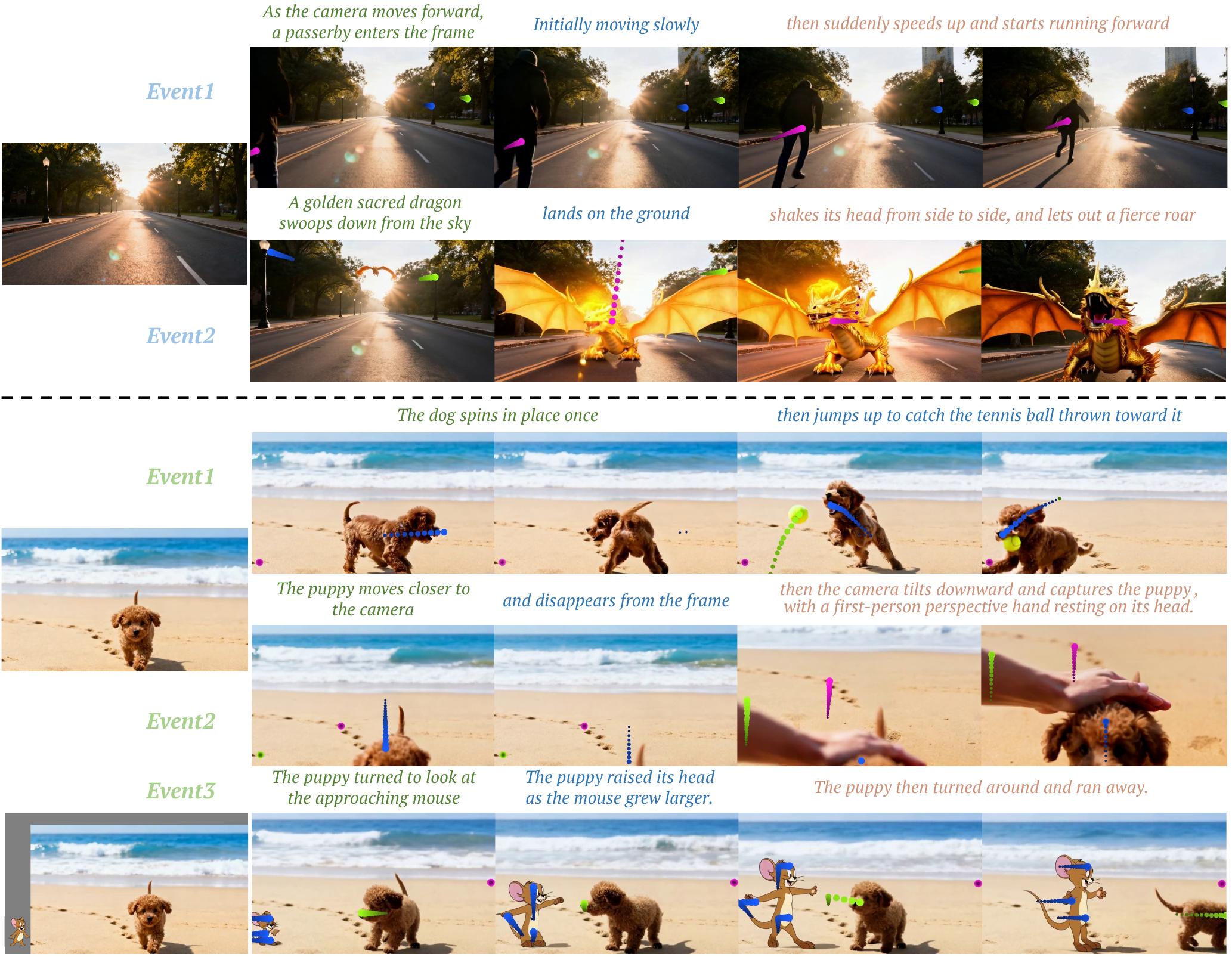}
      \captionsetup{type=figure}
      \caption{By combining text prompts, user-defined trajectories and reference images, our method effectively enables the controllable generation of promptable world events. We strongly recommend viewing our video results in our \href{https://worldcanvas.github.io/}{project page}.}
      \label{fig:teaser}
    \end{center}
}]
\let\thefootnote\relax\footnotetext{\noindent$^\dagger$Corresponding author.}
\maketitle
\input{sec/0_abstract}    
\input{sec/1_intro}
\input{sec/2_related}
\input{sec/3_method}
\input{sec/4_exp}
\input{sec/5_conclusion}
\input{sec/X_suppl}
\newpage
\input{sec/ref.tex}

\end{document}

%% file: sec/0_abstract.tex
\begin{abstract}

We present \method, a framework for promptable world events that enables rich, user-directed simulation by combining text, trajectories, and reference images. Unlike text-only approaches and existing trajectory-controlled image-to-video methods, our multimodal approach combines trajectories---encoding motion, timing, and visibility---with natural language for semantic intent and reference images for visual grounding of object identity, enabling the generation of coherent, controllable events that include multi-agent interactions, object entry/exit, reference-guided appearance and counterintuitive events. The resulting videos demonstrate not only temporal coherence but also emergent consistency, preserving object identity and scene despite temporary disappearance. By supporting expressive world events generation, \method advances world models from passive predictors to interactive, user-shaped simulators. Our project page is available at:
\textcolor{magenta}{\href{https://worldcanvas.github.io/}{https://worldcanvas.github.io/}}.
\end{abstract}

%% file: sec/1_intro.tex
\section{Introduction}
\label{sec:intro}

World models~\cite{matrixgame, matrixgame2, worldmirror, hunyuanworld, genie3, wiedemer2025video}  are unlocking their true potential, evolving from passive simulators into interactive canvases for creation. A landmark step in this evolution is the introduction of ``\textbf{\textit{promptable world events}},'' a concept pioneered by models like Genie~3~\cite{genie3}, which transforms the world model into an interactive canvas where text prompts can trigger significant environmental changes. 

We argue that realizing the full potential of promptable world events requires moving beyond text instructions. Text by itself is ambiguous or insufficient for specifying the complex spatiotemporal dynamics inherent in such events. To enable richer and more fine-grained control, we introduce \method, a framework designed to paint promptable world events with comprehensive control over their fundamental components: \textit{when, where, who,} and \textit{what}.

We decompose the specification of a promptable world event into three complementary modalities:

\begin{itemize}
    \item \textbf{Trajectories} provide the \textit{when} and \textit{where}. They offer a minimal yet powerful interface to encode motion dynamics: their positions define spatial paths, the spacing between points encodes speed, and visibility flags can signal occlusion or entry/exit.
    
    \item \textbf{Reference images} specify the \textit{who}. It provides crucial visual grounding by defining the appearance and identity of the objects or agents involved in the event.
    
    \item \textbf{Text} describes the \textit{what}. It supplies the high-level semantic intent, describing the interactions, goals, and causal structures that constitute the event's narrative.
\end{itemize}
Together, this multimodal triplet forms a complete and unambiguous specification, empowering users to create controllable, promptable world events directly onto the canvas.

A basic solution for promptable events is to use image-to-video generation models conditioned on text or trajectory~\cite{zhang2023i2vgen, blattmann2023stable, Levitor, trailblazer, DragNUWA, DragAnything, Tora, to2, frame-in-out, ATI, geng2025motion, shen2025identity, liu2024physgen, ouyang2024codef, mao2025osv, zhang2025packing}. However, we identify three key shortcomings of existing methods: \textbf{(i) Global vs. local textual control.} Existing approaches typically rely on a single, global text prompt to describe an entire video. This paradigm is insufficient for specifying distinct actions for multiple agents or complex motion dynamics, as it lacks a mechanism to associate specific textual descriptions with individual trajectories. Our work addresses this by establishing a direct correspondence between motion-focused text and its corresponding trajectory, enabling fine-grained, localized control. \textbf{(ii) Lossy representation of trajectories.} Current methods treat trajectories as a mere sequence of spatial coordinates, discarding the wealth of spatiotemporal information they implicitly contain. Critical details like the timing of actions, speed variations encoded in point spacing, and visibility flags for occlusion or scene entry/exit are lost. 
\textbf{(iii) Reference-based control is incomplete.} Current I2V methods lack a robust mechanism to intuitively support reference-based generation, making it difficult for users to directly integrate their reference images with the video to control the subject of the generated event. Consequently, these models fail to generate complex, compositional events.

To bridge these gaps, we propose \method, a framework for promptable world event generation that redefines the synergy between trajectory, text and reference images. Specifically, we curate a dataset of trajectory–video–text triplets with action-centric captions tightly aligned to motion cues. We then employ data augmentation techniques like cropping 
to train our model’s capability in handling objects entry/exit and reference-based generation. Using this data, we adapt Wan 2.2 I2V~\cite{Wan} by injecting trajectory information
and introducing a spatial-aware cross-attention mechanism to bind each textual phrase to its corresponding trajectory region---critical for multi-agent scenarios. During inference, we design an interface that allows users to conveniently input trajectories and reference images, thereby fully leveraging their semantic, spatiotemporal, and appearance information.
Built upon these capabilities, our method can be readily integrated into world models to effectively support user-controllable, semantically rich, and highly interactive promptable world events.

Our experiments show that \method enables users to generate complex, coherent events with fine-grained control while maintaining consistency---even when objects temporarily leave and re-enter the scene. These capabilities suggest that our approach captures structured world dynamics beyond surface-level generation. By providing a practical interface for specifying and simulating user-defined events, \method represents a step toward interactive, controllable world models—ones that don’t just predict the future, but let users shape it.

%% file: sec/2_related.tex
\section{Related Work}
\label{sec:related}
\subsection{Promptable World Events}
World models~\cite{agarwal2025cosmos, matrixgame, matrixgame2, worldmirror, hunyuanworld, genie3, hafner2023mastering, wu2025video, ren2025videoworld, huang2025vid2world, po2025long, kang2024far, he2025cameractrl} aim to learn internal representations of environmental dynamics to support prediction and generation. However, existing approaches predominantly focus on low-level reconstruction, with user control limited to passive observation such as key-based navigation or camera pose control, rather than active changes in the environment. Crucially, these models largely overlook the generation of world events: semantically meaningful environmental changes that unfold over time. Recently, Genie3~\cite{genie3} introduced the concept of ``promptable world events," enabling users to specify high-level events via textual prompts and thereby simulate richer, more interactive scene dynamics---an essential step toward training agents in complex, open-world settings. Yet, text-only prompts often lack the spatial, temporal, and behavioral precision required to control complex, compositional events. To address this limitation, we extend the notion of promptable world events through a multimodal prompting paradigm that integrates textual instructions with reference images and trajectories. This enables fine-grained specification of \textit{where, when, what} happens, and \textit{who} is involved, significantly enhancing controllability and paving the way toward truly interactive and generative world models.

\subsection{Trajectory-Controlled Video Generation}
Trajectory-controlled I2V generation~\cite{namekata2024sg, Levitor, trailblazer, DragNUWA, DragAnything, Tora, to2, frame-in-out, ATI, shi2024motion, lei2025animateanything, gillman2025force, yariv2025through, chuwan} represents an instantiation of such multimodal prompting. However, existing approaches suffer from key limitations that hinder their ability to model complex world events. On the one hand, textual captions are usually global and fail to capture motion or localized semantics, while trajectories serve only as coarse positional cues, ignoring their visibility and spatiotemporal content. On the other hand, most methods offer limited customizability. Although Frame-In-Out~\cite{frame-in-out} enables some customization, its architecture and lack of joint text-trajectory-image modeling limit its capacity for event-level control. In contrast, we introduce a pipeline for curating high-quality multimodal prompting data, coupled with a Spatial-Aware Weighted Cross-Attention mechanism that effectively fuses reference images, trajectory sketches, and detailed textual instructions into the powerful pretrained Wan~\cite{Wan} model, enabling precise, compositional control over world events.

%% file: sec/3_method.tex
\begin{figure*}[t] 
    \centering
    \includegraphics[width=\textwidth]{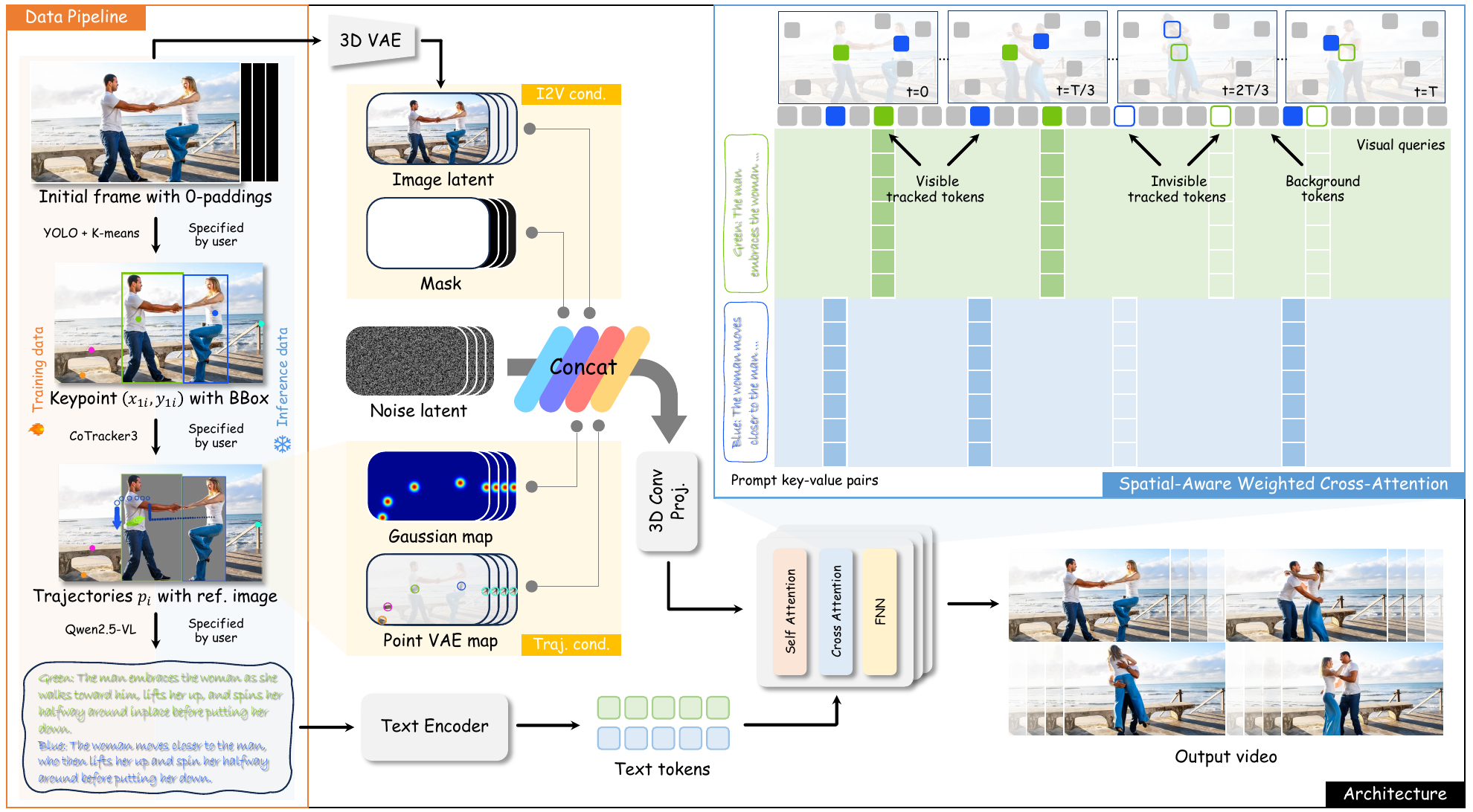} 
    \caption{\textbf{The architecture of our \method.} The data pipeline generates high-quality trajectory–reference–text triplets (in the figure, gray boxes denote reference images extracted from the video, and hollow circles along trajectories indicate invisible points due to occlusion or rotation). The Spatial-Aware Weighted Cross-Attention mechanism explicitly aligns each caption with its associated trajectory.} 
    \vspace{-3ex} 
    \label{fig:overview} 
\end{figure*}

\section{Method}
\label{sec:method}
We build \method based on Wan2.2 I2V 14B model~\cite{Wan}.
In the following sections, we detail our data curation pipeline (\cref{data}), model architecture (\cref{model}), training objective and inference system (\cref{train_infer}).
The overall architecture of \method is shown in ~\cref{fig:overview}.

\subsection{Data Curation Pipeline}
\label{data}
Previous video annotations primarily consist of trajectory labels and global captions, which fail to meet our trajectory–reference–text triplet requirements. In contrast, our pipeline produces annotations that:
(i) explicitly establish correspondence between individual trajectories and captions, ensuring accurate semantic–spatiotemporal alignment in multi-agent scenes or complex motion dynamics;
(ii) effectively extract reference images, enabling precise visual grounding of textual descriptions.
Our data curation pipeline includes the following components:

\noindent \textbf{Keypoints tracking and filtering:} We begin by collecting a diverse set of videos from publicly available sources and segment them into shot-consistent clips with a shot boundary detection algorithm~\cite{transnetV2}. For each clip, we detect objects in the first frame using YOLO~\cite{yolo} to obtain semantically meaningful bounding boxes corresponding to foreground objects. Leveraging these bounding boxes as prompts, we generate corresponding masks with SAM~\cite{sam}, and select 1–3 representative keypoints per mask via K-means. Trajectories of these keypoints will be the primary focus of our subsequent analysis. Additional keypoints are randomly sampled outside foreground object bounding boxes to capture background motion. Once the initial set of keypoints is established, we track them throughout the entire clip using CoTracker3~\cite{cotracker3}, yielding trajectories and visibility scores. After that, we apply random cropping to video frames, which can result in tracked foreground objects being absent in the first frame and reappearing later as they move back into view, effectively simulating objects entering the scene from outside. To prioritize clips containing semantically rich and visually dynamic motion, we further compute a motion score as the average cumulative displacement of all tracked points and filter out clips below a threshold, discarding near-static sequences and preserving high-quality, semantically meaningful tracking data.

\noindent \textbf{Trajectory-based motion captioning:} To generate motion-focused captions rather than generic appearance-based global descriptions, we first visualize only the foreground object trajectories on the original video, with all trajectories from the same object rendered in a consistent color to preserve identity. We then use Qwen2.5-VL 72B model~\cite{qwenvl} to caption these trajectory-visualized videos, prompting the model to identify each colored trajectory’s subject and describe its motion in detail. To ensure the resulting captions remain focused on motion, we retain only minimal subject identifiers (e.g., “a man”) and combine them with the detailed motion descriptions to form the final caption. By using color as an intermediary bridge between visual trajectories and language, this approach yields motion-rich captions while establishing a clear correspondence between each subject and its associated trajectory.

\noindent \textbf{Reference images extraction:} We treat the foreground objects detected by YOLO in the first frame of the video as motion agents and apply mild affine transformations (e.g., translation, scaling, rotation) to these objects to generate reference images. This approach removes constraints on the number and initial positions of reference objects, enabling support for an arbitrary number of reference images starting motion from any location. Moreover, using such transformed data compels the model to achieve modality alignment between the reference image and the video, allowing users to intuitively control the reference image’s initial position, scale, and pose through a simple ``drag-and-drop" interface to animate it. Combined with random cropping, this also enables simulation of scenarios where a user-provided object (defined by a reference image) enters the scene from off-screen, thereby supporting more flexible and interactive generation of promptable world events.

\subsection{Video Generation with Multimodal Triplet}
\label{model}
Using our data curation pipeline, we can get our desired multimodal triplet for distinct objects in the video, represented as:
\begin{equation}
    \mathcal{T} = \{(p_{i}, bbox_i, cap_i)\}_{i=1}^N,
\end{equation}
where $N$ denotes the number of foreground object trajectories, $p_{i} = \{(x_{ti}, y_{ti}, v_{ti})\}_{t=1}^T$ denotes the 2D coordinates and visibility score of the $i_{th}$ trajectory. $bbox_i$ 
is the bounding box of associated foreground object in the initial frame, and $cap_i$ is the motion-focused textual description. For background points, only 2D coordinates and visibility are available. Note that for brevity, we omit explicit mention of reference images here and assume they have already been inserted into the first video frame as image condition (when reference images are provided).
In this section, we present our core model design and its use of this multimodal triplet for controllable video generation. Our approach consists of two key components: Trajectory Injection for motion control via input trajectories, and Spatial-Aware Weighted Cross-Attention to align trajectories with text prompts in multi-agent scenarios.
\subsubsection{Trajectory Injection}
We introduce additional conditional feature channels to inject trajectory information as a control signal during training, enabling the model to generate motion guided by user-specified trajectories. Specifically, we represent all trajectories using Gaussian heatmap and
propagate the image VAE latent feature at the first-frame keypoint location $(x_{1i}, y_{1i})$ of each trajectory to all subsequent points along that trajectory—forming what we call a point VAE map—to enhance the model’s ability to follow and animate specific points over time. The Gaussian heatmap and point VAE map are then concatenated directly with the original DiT inputs of Wan2.2 I2V model (noise latent, conditional image latent and rearranged mask) along the channel dimension. Before being passed through the Wan DiT model, the combined features are processed by a 3D convolutional layer to align their dimensionality with that of the DiT blocks, thereby seamlessly integrating the trajectory-based control signals into the generative process. Weights for the newly added Gaussian heatmap and point VAE map in the 3D convolutional layer are initialized with zero values.

\subsubsection{Spatial-Aware Weighted Cross-Attention}
In multi-agent scenarios, especially when subjects look similar or lack distinct visual cues, matching multiple captions to their corresponding trajectories is challenging. To address this, we propose Spatial-Aware Weighted Cross-Attention that explicitly aligns each caption with its associated trajectory.

Instead of computing cross-attention between all video tokens and the full text prompt uniformly, our method encourages the model to focus more on visual tokens that spatially overlap with each trajectory. Specifically, for each data triplet $(p_{i}, bbox_i, cap_i)$, we use the size of the bounding box $bbox_i$ as the spatial extent of trajectory $i$. Then, the coverage area of trajectory $i$ is the region centered at $(x_{ti}, y_{ti})$ with the same width and height as $bbox_i$. Denote the visual query tokens that fall within this coverage area as $Q_i$, the corresponding key-value pair ${KV_i}$ is derived from the text embedding of ${cap_i}$.
During cross-attention, we assign a higher weight bias to the attention scores between $Q_i$ and $KV_i$. This operation can be expressed as:
\begin{equation}
W_{qk} = 
\begin{cases}
log(w), & {\rm if} \;\; v_{ti} =1 \text{ and } Q_q \in Q_i \text{ and } K_k \in K_i, \\
0, & {\rm otherwise}.
\end{cases}
\end{equation}
Here, $w$ is empirically set as $30$. After computing the weight aggregation over all trajectories, we apply the resulting weight matrix to the cross-attention computation:
\begin{equation}
\text{Attention}(Q, K, V) = \text{Softmax}(\frac{QK^T}{\sqrt{D}} + W) V.
\end{equation}
Here, $W$ gives larger values to token pairs that belong to the same trajectory-caption pair. This design strengthens the correspondence between trajectories and their descriptive text, while still allowing the model to attend to other regions when needed.

\subsection{Training and Inference}
\label{train_infer}
We follow the flow matching framework~\cite{flow, flow2} to perform post-training using $L1$ reconstruction loss. Given random noise $x_0 \sim \mathcal{N}(0, I)$, timestep $t \in [0, 1]$ and ground-truth video latent $x_1$, we calculate the training input $x_t = tx_1 + (1-t)x_0$, ground truth velocity $v_t = \frac{dx_t}{dt}=x_1-x_0$, the training objective is formulated as:
\begin{equation}
\mathcal{L}=\mathbb{E}_{x_0, x_1, t, \mathcal{C}}\left[\left\|u(x_t,t,\mathcal{C;\theta})-v_t\right\|^2\right],
\label{eq:conditonal_denoising}
\end{equation}
where $\mathcal{C}$ is the union of all conditions, $\theta$ is model parameters and $u(x_t,t,\mathcal{C;\theta})$ denotes the output velocity predicted by the model.

During inference, we provide an intuitive interface to specify desired inputs through the following designs: 

\begin{itemize}[noitemsep,topsep=3pt]
\item
\noindent \textbf{Trajectory timing}: Users can define the start and end times of a trajectory, controlling when an object begins and stops moving or when it enters or exits the frame.

\item
\noindent \textbf{Trajectory shape via point sequences}: Trajectories are defined by a sequence of points, with equal time intervals between consecutive points. The spacing between points implicitly controls motion speed: sparser points for faster movement, while denser points for slower motion.

\item
\noindent \textbf{Visibility control}: Users can freely specify which segments of a trajectory are visible or not, enabling realistic modeling of occlusions and complex semantic actions.

\item
\noindent \textbf{Text-trajectory binding}: Each trajectory is associated with a motion caption, ensuring semantic alignment between the described action and the controlled motion.

\item
\noindent \textbf{Reference image insertion}: Users can place various reference images on the canvas and adjust their position and size to introduce specified subjects.
\end{itemize}

These designs help users to create rich, controllable, and semantically meaningful events through intuitive trajectories, text inputs and reference images.

%% file: sec/4_exp.tex
\begin{figure*}[t] 
    \centering
    \includegraphics[width=0.96\textwidth]{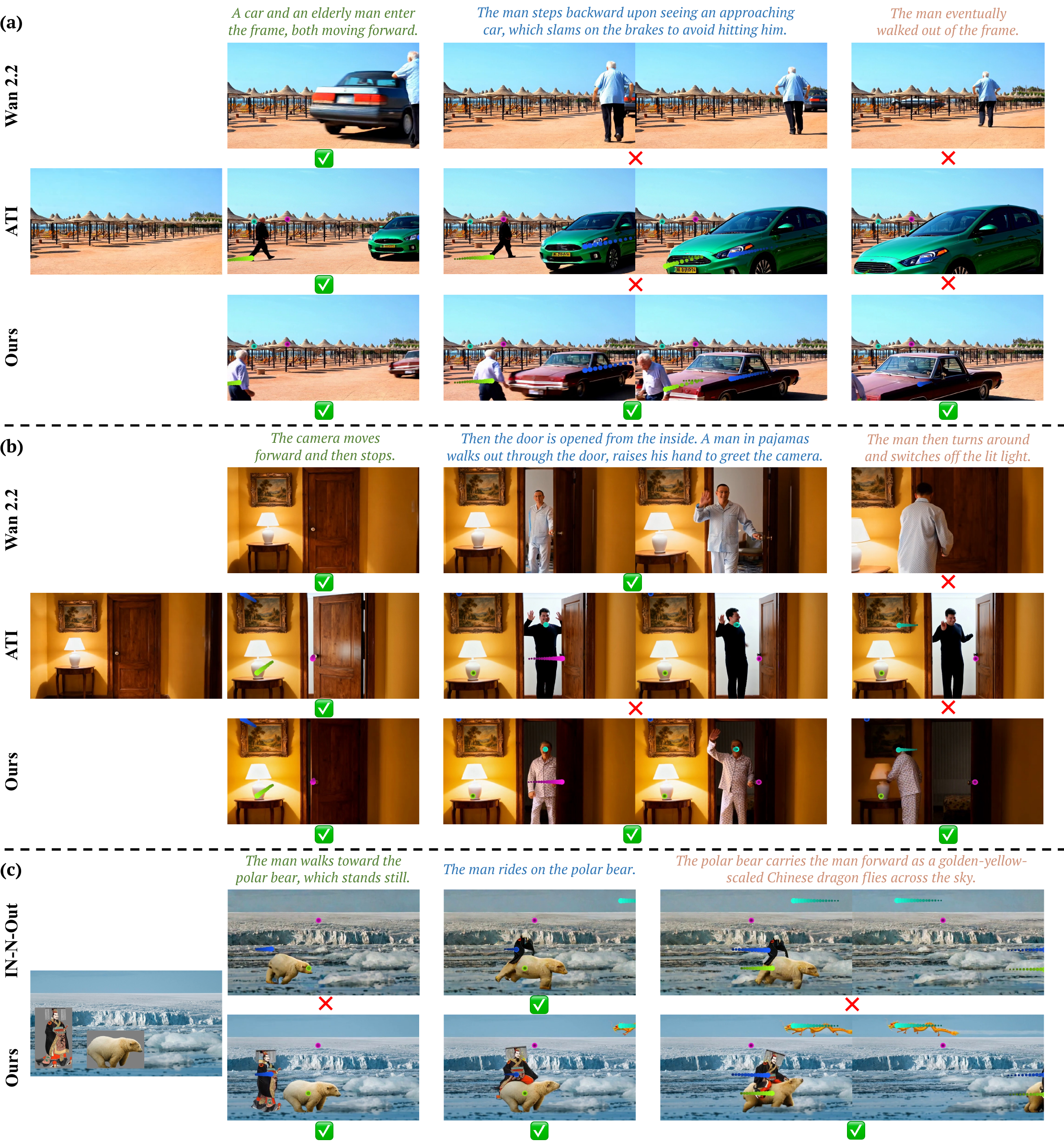} 
    \caption{\textbf{Qualitative comparison on promptable world event modeling.} Our model successfully generates results that align with given trajectories, text prompt and reference images, whereas the baselines fail to properly correspond to these inputs.} 
    \vspace{-2ex} 
    \label{fig:comparison} 
\end{figure*}

\input{tabs/quant_compare}

\begin{figure}[t]
    \centering
    \includegraphics[width=\columnwidth]{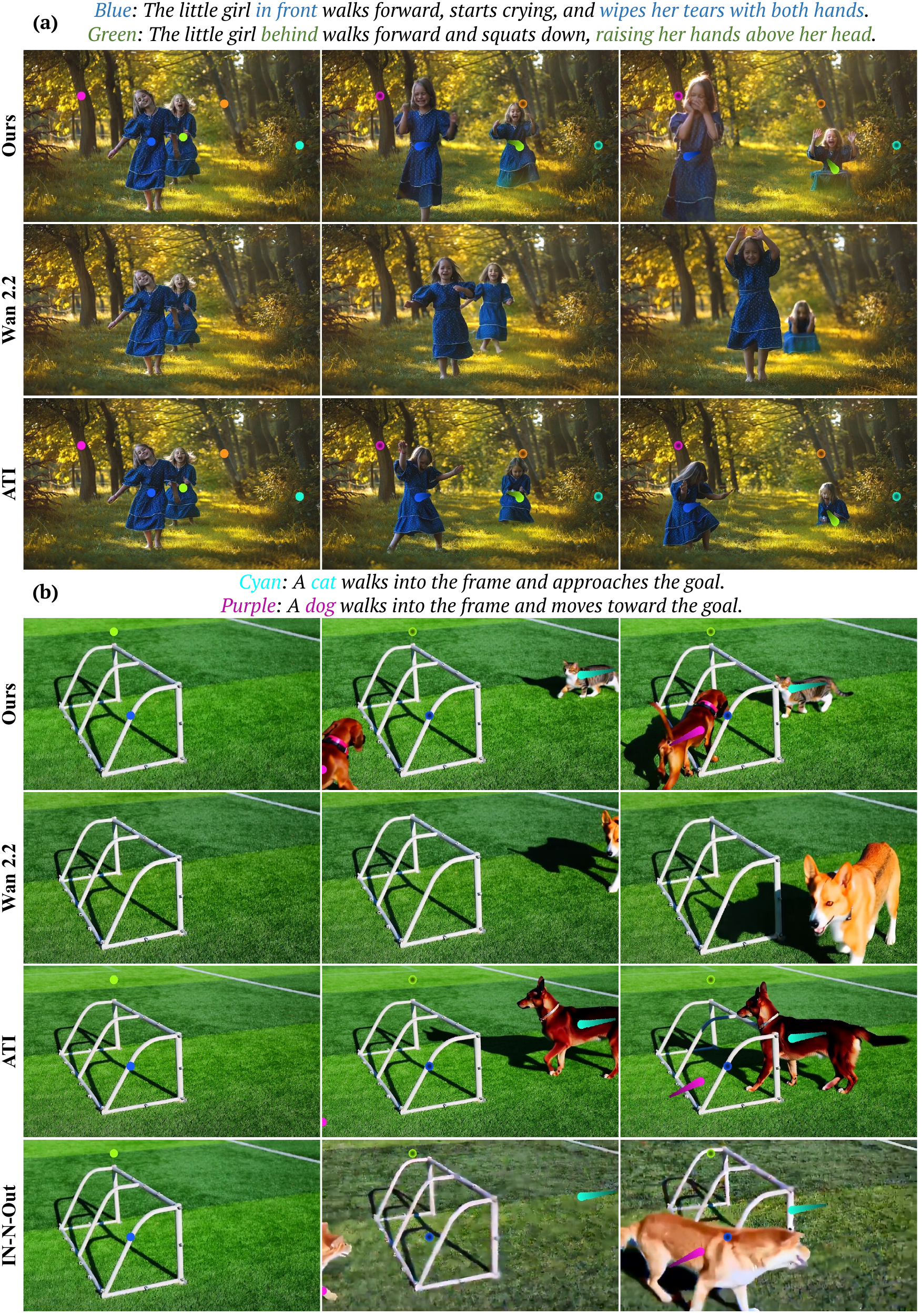}
    \caption{\textbf{Qualitative comparison of multi-subject trajectory-text alignment.} Our method accurately aligns the textual descriptions with motions specified by trajectories, whereas the baselines fail to produce correct results in such cases.}
    \label{fig:corr}
    \vspace{-10pt}
\end{figure}
\section{Experiments}
\label{sec:exp}
In this section, we demonstrate how \method addresses key limitations of prior models in supporting promptable world events. 

Specifically, ~\cref{imple} details the implementation of our model. In ~\cref{ssec:comparison}, we compare \method with state-of-the-art methods and demonstrate our superior performance over these baselines. Next, ~\cref{advance} demonstrates the advanced consistency maintenance capability of \method. Finally, we validate the effectiveness of the Spatial-Aware Weighted Cross-Attention mechanism through an ablation study in ~\cref{ablation}.

\subsection{Implementation Details}
\label{imple}
We train our model on our curated dataset of 280k trajectory–video–text triplets. All videos are processed at a resolution of $480 \times832$. Our model is trained for 9k steps with a learning rate of $1 \times 10^{-5}$ and a linear warmup schedule. The entire training process is conducted on 64 NVIDIA H800 GPUs, with a total batch size of 64.

\subsection{Comparison}
\label{ssec:comparison}
We select the powerful base model Wan2.2 I2V 14B~\cite{Wan}, ATI~\cite{ATI} (currently the strongest open-source I2V model), and Frame In-N-Out~\cite{frame-in-out} (an I2V model supporting reference-guided generation) as our baselines, to evaluate our model’s capabilities in semantic understanding (\textit{what}), trajectory following (\textit{when} and \textit{where}), and reference-based generation (\textit{who}), respectively.

\subsubsection{Qualitative Results}

~\cref{fig:comparison} shows our leading results across \textit{what, when, where}, and \textit{who}, demonstrating its strong capability in generating promptable world events. In the tested event generation examples, both Wan and ATI fail to accurately understand the intended event. For instance, in example (a), Wan generates an elderly man chasing a car, while ATI neither follows the trajectory of the man nor produces the expected interaction between the old man and the car. In example (b), neither Wan nor ATI succeeded in turning off the light. These results highlight that accurately generating complex events requires precise coordination between textual understanding and exact trajectory following. Without both, the output diverges significantly from user expectations. Our approach successfully achieves what previous text control methods and I2V models could not by correctly understanding both the text and the trajectory while ensuring their correspondence, enabling faithful and controllable generation of world events. In the reference-based generation example (c), Frame In-N-Out fails to adequately preserve the consistency of the reference images and also misinterprets the intended event. In contrast, our model maintains identity consistency with the reference images, treating them as the central subjects of the event. By seamlessly integrating the reference images with motion trajectories and text prompt, our approach accurately generates the desired output.

We also conduct comparisons on Trajectory-Text Alignment capability, which is essential for generating coherent multi-subject events. This requires the model to correctly associate each caption with its intended agent, especially when multiple entities interact or appear dynamically.
~\cref{fig:corr} evaluates this capability on two scenarios: controlling existing subjects in the scene and multiple objects entering from off-screen. In case (a), Wan partially captures subject distinctions from the text prompt (``girl in front" and ``girl behind") but still fails to fully respect the described actions. For instance, the front girl raises her hand despite the prompt specifying otherwise. ATI, limited by weaker text grounding, misinterprets both children’s motions. In the more challenging case (b), all baselines merely generate a dog entering the frame, failing to capture the full semantic intent of the text prompt. In contrast, our model correctly binds each trajectory to its caption and generates the complete, intended multi-agent events.

\subsubsection{Quantitative Results}
We evaluate our model from two complementary perspectives: trajectory following accuracy and semantic understanding. 
Specifically, we collect 100 image–trajectory pairs depicting semantic events and generate videos using our model. For each output, we apply CoTracker3~\cite{cotracker3} to track the user-specified points and get the generated trajectories. We report the following metrics: ObjMC~\cite{motionctrl}: the mean Euclidean distance between generated and user-defined trajectories; Appearance Rate~\cite{ATI}: the proportion of frames in which the tracker accurately predicts a point as visible whenever the input trajectory indicates it is visible; Subject \& Background Consistency~\cite{vbench}: metrics reflect temporal consistency in video generation;  CLIP-T~\cite{tgt}: score reflecting semantic alignment between text and generated video, validated at both global and local levels.
Results in ~\cref{tab:evaluation} show that our method consistently outperforms the baseline across three aspects: trajectory following, semantic alignment, and generated video quality, demonstrating its superiority in supporting promptable world events generation.
More details on these metrics can be found in the supplementary material.

\begin{figure*}[t] 
    \centering
    \includegraphics[width=0.95\textwidth]{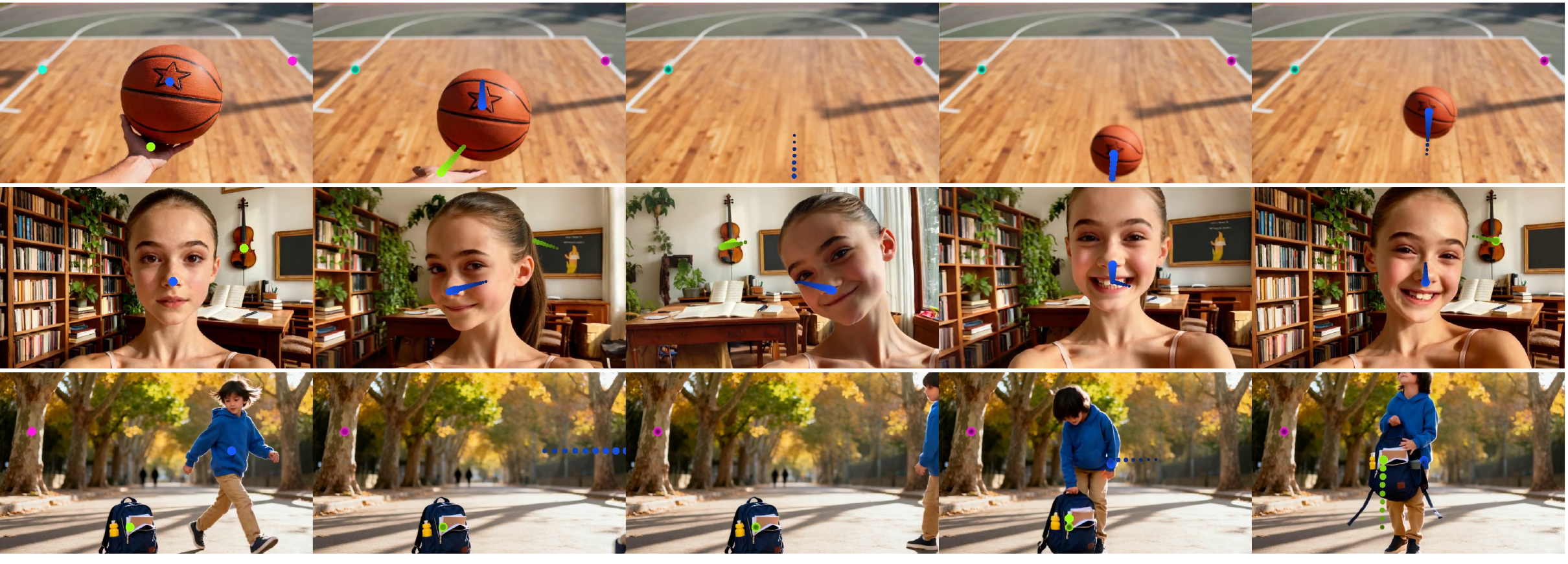} 
    \caption{\textbf{Consistency maintenance results.} The shown examples correspond to object consistency preservation, scene consistency preservation, and character consistency preservation, respectively.} 
    \vspace{-2ex} 
    \label{fig:memory} 
\end{figure*}

\subsection{Consistency Maintenance Capability}
\label{advance}

We also observe that \method exhibits a strong ability to maintain scene and object consistency over time, even when subjects temporarily leave the frame and reappear later. This temporal coherence ensures that objects retain their appearance, identity, and spatial relationships across occlusions or off-screen intervals.
~\cref{fig:memory} illustrates such cases, where characters or scenes exit the view and re-enter while preserving visual and semantic consistency. This capability is crucial for generating plausible world events and reflects an emergent form of visual memory in the model. It not only enhances realism but also highlights our model’s potential as a building block for more advanced world models capable of coherent, persistent scene simulation.

\begin{figure}[t]
    \centering
    \includegraphics[width=\columnwidth]{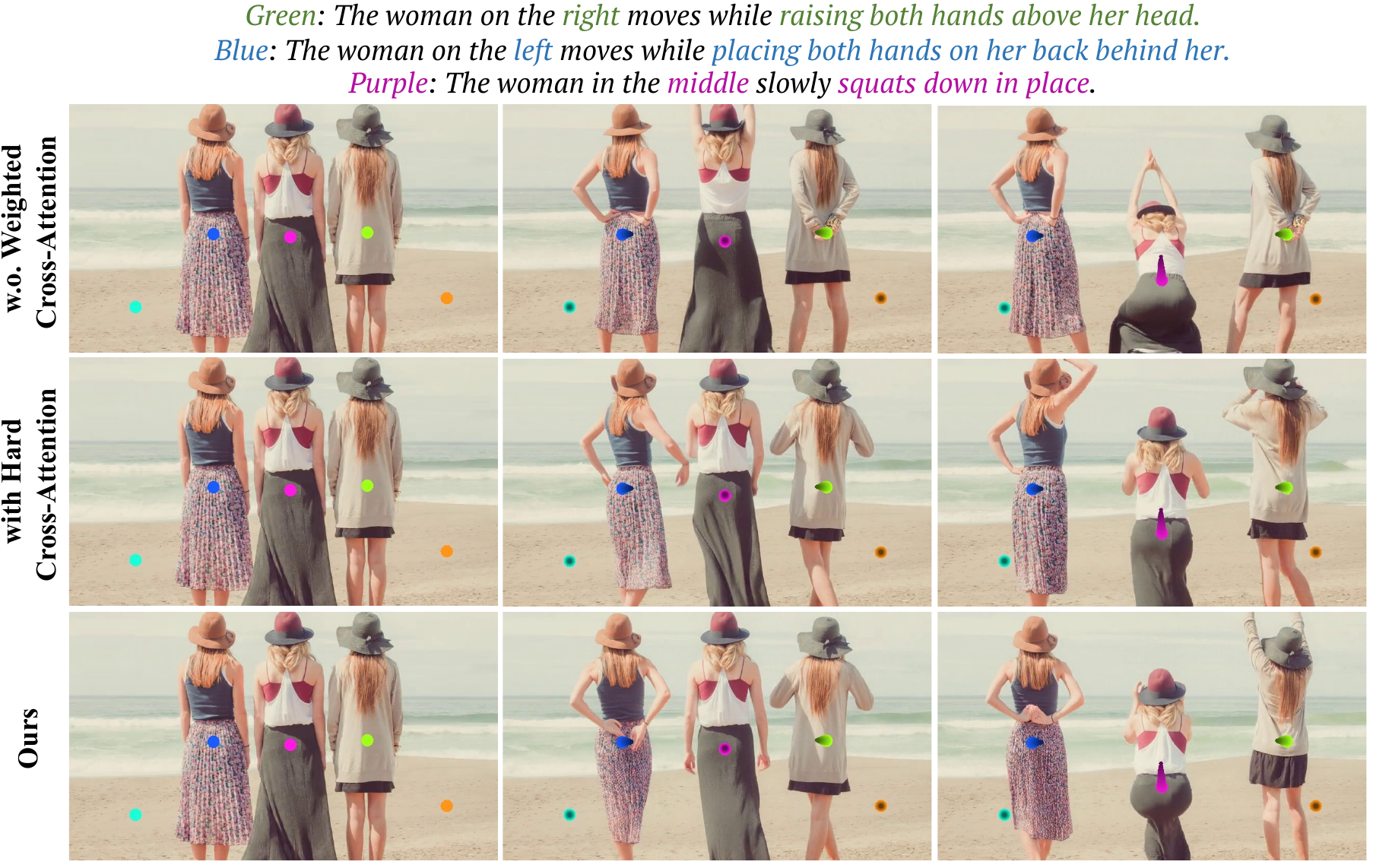}
    \caption{\textbf{Qualitative results for our ablation study.} Compared to variants \textit{without Spatial-Aware Weighted Cross-Attention} and \textit{Hard Cross-Attention}, the former causes severe semantic-action misalignment, while the latter yields incomplete semantics. In contrast, our method effectively achieves accurate alignment between semantic content and trajectories.}
    \label{fig:ablation}
    \vspace{-15pt}
\end{figure}

\subsection{Ablation Study}
\label{ablation}
We conduct ablation experiments to validate the effectiveness of our Spatial-Aware Weighted Cross-Attention mechanism. ~\cref{fig:ablation} shows a qualitative comparison on multi-subject trajectory–text alignment: when using standard full cross-attention (without spatial weighting) or hard cross-attention (only keep attention weights in the region of interest, and set all other positions to zero), the model fails to associate captions with their correct trajectories, leading to swapped or mismatched actions. In contrast, our spatial-aware variant successfully preserves the intended correspondences.
Quantitative results in the supplement further confirm this trend: our method achieves higher alignment accuracy and video quality, demonstrating that the spatial weighting is critical for correct trajectory–text binding in complex multi-agent scenarios.

%% file: tabs/quant_compare.tex
\begin{table*}[tb]
  \scriptsize
  \caption{\textbf{Quantitative results}. The best and runner-up are in \textbf{bold} and \underline{underlined}.}
  \label{tab:evaluation}
  \centering
  \resizebox{\textwidth}{!}{%
  \renewcommand{\arraystretch}{1.2}%
  \begin{tabular}{p{2.0cm}cccccccccc} %

\toprule
\multirow{2}{*}{Method} 
    & \multirow{2}{*}{\parbox{1.5cm}{\centering \textbf{ObjMC}\textdownarrow}}
    & \multirow{2}{*}{\parbox{1.8cm}{\centering \textbf{Appearance Rate}\textuparrow}}
    & \multirow{2}{*}{\parbox{1.8cm}{\centering \textbf{Subject Consistency}\textuparrow}} 
    & \multirow{2}{*}{\parbox{1.8cm}{\centering \textbf{Background Consistency}\textuparrow}} 
    & \multirow{2}{*}{\parbox{1.5cm}{\centering \textbf{CLIP-T\\(Global)}\textuparrow}} 
    & \multirow{2}{*}{\parbox{1.5cm}{\centering \textbf{CLIP-T\\(Local)}\textuparrow}} 
\\
\\ 
    \hline
    WAN2.2 I2V & 139.59 & 70.65\% & \underline{0.8947} & 0.9192 & 0.1727 & \underline{0.1678} \\
    ATI & \underline{127.21} & \underline{80.44}\% & 0.8850 & \underline{0.9225} & 0.1617 & 0.1629 \\
    Frame In-N-Out & 142.70 & 64.74\% & 0.8411 & 0.8852 & \underline{0.1738} & 0.1656 \\
    \hline
    Ours & \textbf{91.06} & \textbf{85.17\%} & \textbf{0.9044} & \textbf{0.9326} &  \textbf{0.1742} & \textbf{0.1680} \\
   \bottomrule
\end{tabular}%
}
\end{table*}

%% file: sec/5_conclusion.tex
\section{Conclusion}
\label{sec:conclusion}
We introduce a trajectory-text-reference-driven paradigm for generating promptable world events---a crucial step toward controllable, semantically grounded video simulation. By enabling users to precisely specify \textit{what, when, where}, and \textit{who} through intuitive motion trajectories, natural language and ref images, our approach supports semantic actions, complex interactions, object entry/exit and reference-guided appearance. Critically, the model exhibits emergent object consistency and scene memory, suggesting it captures structured world dynamics beyond surface-level generation. These capabilities position our framework as a practical and scalable pathway toward interactive world models that can simulate user-defined events with both realism and reliability.

\noindent\textbf{Acknowledgements:} The authors thank \href{https://pixabay.com/}{Pixabay} for providing some images used in this paper under the Pixabay License, which permits free use without attribution.

%% file: sec/X_suppl.tex
\appendix
\renewcommand\thesection{\Alph{section}}
\renewcommand\thefigure{S\arabic{figure}}
\renewcommand\thetable{S\arabic{table}}
\renewcommand\theequation{S\arabic{equation}}
\setcounter{figure}{0}
\setcounter{table}{0}
\setcounter{equation}{0}
\setcounter{page}{1}
\maketitlesupplementary

\section{Details on Quantitative Metrics}
\label{subsec:app_metrics}

We evaluate our model from two complementary perspectives: \emph{trajectory following accuracy} and \emph{semantic understanding}.
Specifically, we construct a benchmark of 100 image–trajectory pairs that describe semantic events and generate corresponding videos using our model. For each generated video, we employ CoTracker3~\citep{cotracker3} to trace the user-specified points and obtain the resulting motion trajectories. The evaluation includes the following metrics:

\begin{itemize}
\item \textbf{ObjMC}~\citep{motionctrl}: Measures trajectory following accuracy by computing the \emph{mean Euclidean distance} between the generated and user-defined trajectories. Lower values indicate more precise motion control.
\item \textbf{Appearance Rate}~\citep{ATI}: Quantifies visibility consistency by calculating the proportion of frames in which the tracker correctly predicts a point as visible whenever the input trajectory marks it as visible.
\item \textbf{Subject \& Background Consistency}~\citep{vbench}: Assesses temporal coherence by evaluating feature-level stability for both the subject and the background across consecutive frames, reflecting smoothness and visual consistency.
\item \textbf{CLIP-T (Global)}~\citep{tgt}: Measures the overall semantic consistency between the generated video and the input text prompt using ViCLIP~\citep{viclip}. Specifically, we compute the cosine similarity between ViCLIP features extracted from the entire video and those from the global text description, reflecting how well the generated scene aligns with the intended semantics.

\item \textbf{CLIP-T (Local)}~\citep{tgt}: Evaluates fine-grained alignment between local text prompts and the corresponding visual regions. Following~\citep{tgt}, we crop local video patches, and compute ViCLIP feature similarities between each cropped region and its associated local text. The final local CLIP-T score is averaged across multiple window sizes, providing a more detailed measure of spatial semantic fidelity.
\end{itemize}

\section{More Quantitative Results}
\label{subsectab:more_quantitative}
In this section, we present additional quantitative experimental results, including quantitative ablation studies of the Spatial-Aware Weighted Cross-Attention mechanism and a user study.

\subsection{Quantitative Ablation Study}
\input{tabs/ablation}

We conducted a quantitative comparison among standard full cross-attention (without spatial weighting), hard cross-attention (which retains attention weights only within the region of interest and sets all other positions to zero), and our proposed Spatial-Aware Weighted Cross-Attention on the evaluation dataset of 100 image–trajectory pairs that we collected. As shown in ~\cref{tab:ablation}, our method achieves the best results across all video quality metrics and semantic alignment metrics, demonstrating the effectiveness of the Spatial-Aware Weighted Cross-Attention mechanism in aligning text and trajectories in complex multi-agent interaction scenarios.
\input{tabs/user_study}

\subsection{Human Evaluation}
In the main text, we quantitatively compare our method against several baselines with metrics in ~\cref{subsec:app_metrics}. However, these metrics have certain limitations. For instance, we observe that the model without trajectory control, Wan2.2 I2V~\cite{Wan}, achieves better ObjMC and Appearance Rate scores than Frame In-N-Out~\cite{frame-in-out}, which does support trajectory control. This counterintuitive result arises because Frame In-N-Out produces outputs with unstable visual quality, leading to inaccurate tracking by CoTracker3~\cite{cotracker3}. Additionally, the CLIP-T metric is not particularly accurate in evaluating fine-grained motion details, resulting in only marginal score differences between different models.

To enable a more precise comparison of the models’ capabilities in terms of following user-specified trajectories, generating semantic events, aligning trajectories with textual descriptions, and preserving reference image information, we conducted a user study for validation.

Specifically, we collected 20 challenging cases involving complex single-object motions, multi-object interactions, and object appearances or disappearances, as well as 10 reference-image-based generation cases (only used for \textit{Reference fidelity} evaluation). These cases were used to generate videos using Wan 2.2 I2V 14B~\cite{Wan}, ATI~\cite{ATI}, Frame In-N-Out~\cite{frame-in-out}, and our \method.
We recruited 15 participants---including video generation researchers, artists, and non-expert users---to ensure a comprehensive and objective evaluation. Each participant was asked to vote for the model they considered best on each of the following five criteria:
\begin{itemize}[nosep, topsep=3pt]
\item \textbf{Trajectory following}: whether the generated video accurately follows the user-provided trajectories;
\item \textbf{Prompt Adherence}: whether the result faithfully reflects the event described in the text prompt;
\item \textbf{Text-trajectory alignment}: whether the content depicted by each trajectory correctly matches its corresponding textual description;
\item \textbf{Reference fidelity}: whether reference-based generations accurately preserve the visual characteristics of the reference image;
\item \textbf{Overall video quality}: whether the generated result is a high-quality video, considering factors such as visual fidelity, motion smoothness, generation consistency, and aesthetics.
\end{itemize}

The results of the user study are shown in ~\cref{tab:user_study}, which clearly demonstrate that \method achieves consistently superior performance across all evaluation metrics, confirming that our model genuinely supports the generation of semantic promptable world events.

\begin{figure*}[t] 
    \centering
    \includegraphics[width=\textwidth]{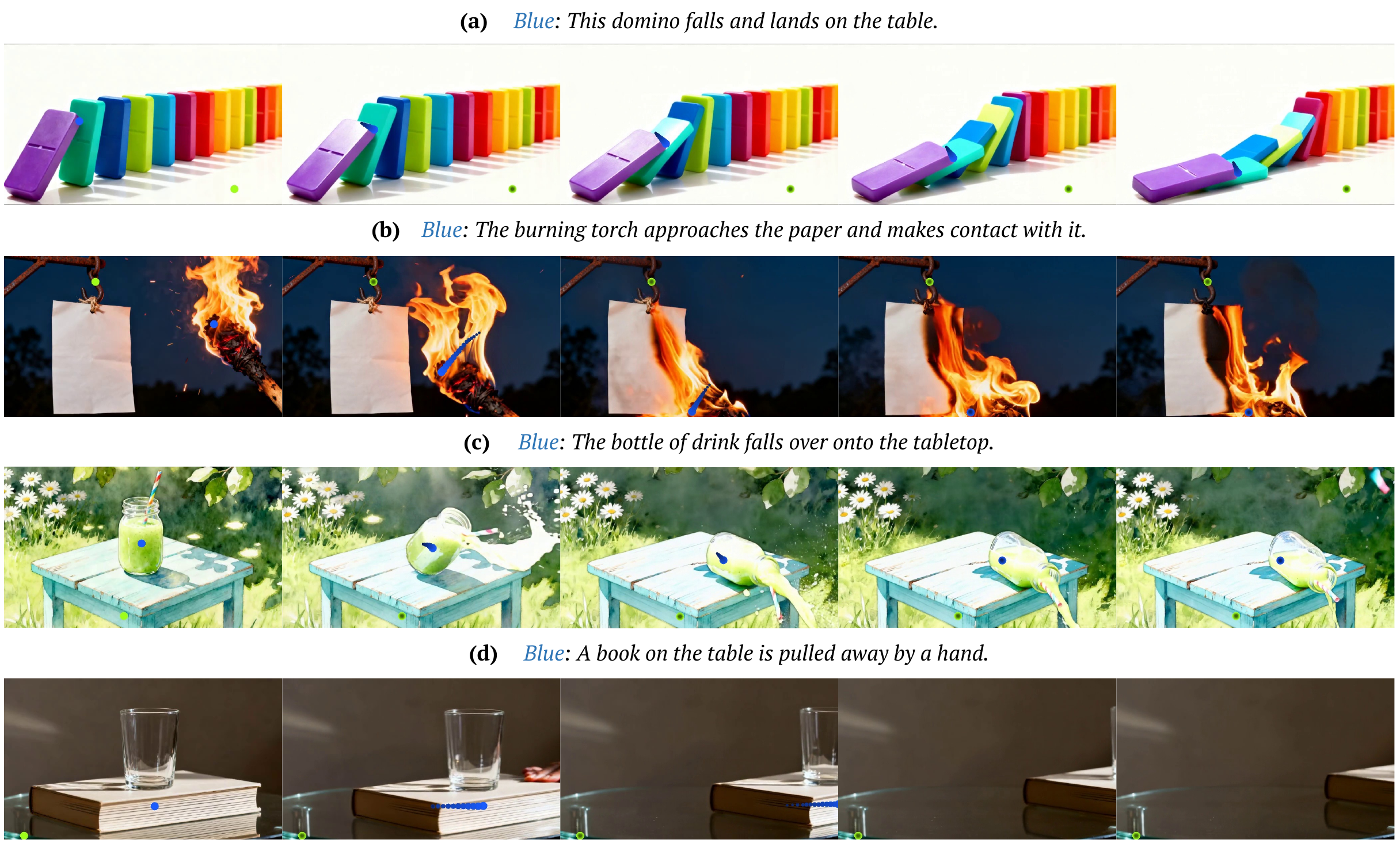} 
    \caption{\textbf{World-model related capabilities.} In these examples, we provided only the trajectories and text describing the "cause," and let the model generate the subsequent outcomes. The results demonstrate that our \method exhibits physical plausibility, causal reasoning, and future prediction abilities. We strongly recommend viewing our video results in our \href{https://worldcanvas.github.io/}{project page}.} 
    \label{fig:supp1} 
\end{figure*}

\begin{figure*}[t] 
    \centering
    \includegraphics[width=\textwidth]{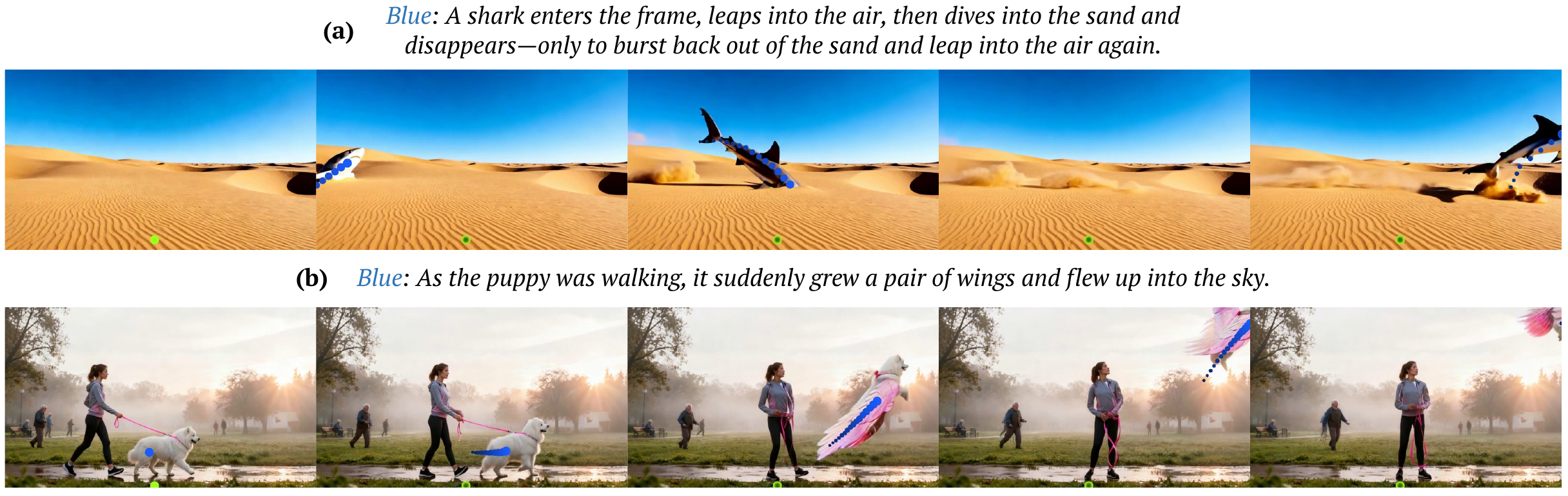} 
    \caption{\textbf{Counterfactual generation results.} Our model is capable of correctly generating counterfactual events that adhere to physical laws and causal logic. We strongly recommend viewing our video results in our \href{https://worldcanvas.github.io/}{project page}.} 
    \label{fig:supp2} 
\end{figure*}

\begin{figure*}[t] 
    \centering
    \includegraphics[width=\textwidth]{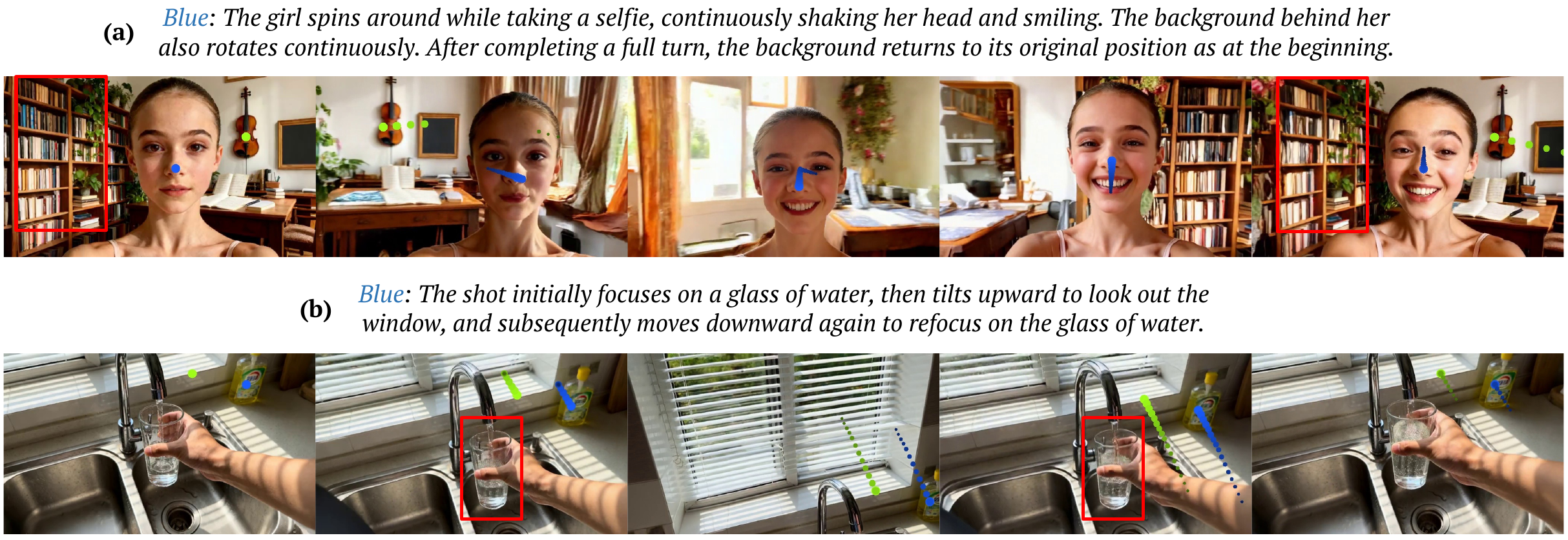} 
    \caption{\textbf{Failure cases.} The red boxes highlight the blurring and distortion in case (a) and the insufficient water level rise in case (b), respectively. We strongly recommend viewing our video results in our \href{https://worldcanvas.github.io/}{project page}.} 
    \vspace{-2ex} 
    \label{fig:supp3} 
\end{figure*}

\section{World-Model Related Capabilities}
In the main text, we highlighted \method’s ability to maintain consistency in objects, scenes, and subjects, ensuring that the content associated with a given trajectory remains temporally coherent throughout. This property is crucial for applying our method to world models. In this section, we explore additional world-model-related capabilities demonstrated by our model, including physical plausibility, causal reasoning, and predictive abilities for future events.

We designed a set of cases to evaluate the world-model related capabilities of \method. In these examples, we only provide trajectories representing the ``cause", while ensuring that the input text prompt describes solely the content of the drawn trajectories and does not mention the resulting effects of such motions. By observing whether \method can correctly generate the subsequent events that naturally follow from the given cause, we can assess its physical plausibility, causal reasoning, and future prediction abilities.

The evaluation results shown in~\cref{fig:supp1} demonstrate \method’s physical plausibility, causal reasoning, and future prediction abilities. Specifically, 
\begin{itemize}[nosep, topsep=3pt]
\item Case (a) depicts a domino chain reaction, where our model accurately generates the sequential falling of dominoes---each tile begins to fall only after being struck by the previous one, rather than all falling simultaneously or remaining upright.
\item Case (b) shows a burning torch approaching a sheet of paper. Although we only control the motion of the torch, the model correctly generates the subsequent charring of the paper as it catches fire.
\item Case (c) shows a drink-filled bottle that is tipped over, and our model realistically renders the liquid spilling out, with the amount of liquid inside the bottle progressively decreasing.
\item In case (d), we only specify the motion of a book being pulled away; nevertheless, the generated video correctly reflects the presence of friction---the glass cup resting on the book moves along with it, and the reflection on the table surface adheres to physical laws.
\end{itemize}
These results collectively confirm that our model exhibits physical plausibility, causal reasoning, and future prediction capabilities, further underscoring \method’s potential for promptable world event generation.

\section{Counterfactual generation}
We further evaluated \method’s ability to generate counterfactual events. This capability provides clearer evidence of the model’s controllability and semantic alignment, as well as its grasp of commonsense knowledge. ~\cref{fig:supp2} presents two generated counterfactual events by \method.
In case (a), a shark leaps out of the desert sand, dives back in, and leaps again. Although this scenario is physically impossible in reality, the model correctly simulates the shark submerging into the sand and the resulting plumes of dust upon each leap.
In case (b), the dog flies upward into the sky, and the model generates the leash snapping while the owner stands on the ground in astonishment.
These counterfactual results confirm that \method indeed understands certain physical principles and causal relationships, and further validate its strong controllable generation capability and semantic fidelity.

\section{Failure cases}
Although our model demonstrates strong performance in generating promptable world events, we observe that it sometimes fails to produce correct results in particularly challenging scenarios involving complex spatial transformations or logical reasoning. ~\cref{fig:supp3} illustrates such failure cases:

In case (a), we instruct the camera to rotate $360$ degrees and return to the original view. While the model largely fulfills this request, the books on the shelf exhibit noticeable blurriness and inconsistencies (highlighted in the red boxes).
Case (b) presents a more complex logical reasoning scenario: initially, as the cup is being filled with water, the water level should rise. When the camera moves upward and the cup is temporarily out of view, the filling process should continue, so the water level must continue to rise. Finally, when the camera refocuses on the cup, the water level should reflect the cumulative filling. The generated result correctly shows the water level rising while the cup is visible, but after the camera pans away and returns, the water level shows no significant increase---contradicting the expected physical behavior.

These examples also highlight key challenges for future research in promptable world event generation: how to ensure consistency under drastic and complex motions, and how to enable the model to perform correct logical reasoning about content that is currently out of view.

%% file: tabs/ablation.tex
\begin{table}[tb]
  \scriptsize
  \caption{\textbf{Quantitative ablation results}. The best is in \textbf{bold}.}
  \label{tab:ablation}
  \centering
  \resizebox{0.48\textwidth}{!}{%
  \renewcommand{\arraystretch}{1.2}%
  \begin{tabular}{p{1cm}cccc} %

    \toprule

    \multirow{2}{*}{Method} 
    & \multirow{2}{*}{\parbox{1.5cm}{\centering \textbf{Subject Consistency}\textuparrow}} 
    & \multirow{2}{*}{\parbox{1.5cm}{\centering \textbf{Background Consistency}\textuparrow}} 
    & \multirow{2}{*}{\parbox{1cm}{\centering \textbf{CLIP-T\\(Global)}\textuparrow}} 
    & \multirow{2}{*}{\parbox{1cm}{\centering \textbf{CLIP-T\\(Local)}\textuparrow}} 
    \\
    \\ 

\hline
    full attn. & 0.8948 & 0.9285 & 0.1694 & 0.1595 \\
    hard attn. & 0.9002 & 0.9277 & 0.1713 & 0.1647 \\
    Ours & \textbf{0.9044} & \textbf{0.9326} &  \textbf{0.1742} & \textbf{0.1680} \\
   \bottomrule
  \end{tabular}%
}
\vspace{-5pt} 
\end{table}

%% file: tabs/user_study.tex
\begin{table*}[tb]
\scriptsize
\caption{\textbf{Human evaluation results}. We report the percentage of ``best'' votes received by each model from our user study across five criteria. The best and runner-up are in \textbf{bold} and \underline{underlined}.}
\label{tab:user_study}
\centering
\renewcommand{\arraystretch}{1.2}
\begin{tabular}{p{3cm}ccccc} 
\toprule
\multirow{2}{*}{Method} & \multirow{2}{*}{\parbox{2cm}{\centering \textbf{Trajectory following}\textuparrow}}
& \multirow{2}{*}{\parbox{1.5cm}{\centering \textbf{Prompt Adherence}\textuparrow}}
& \multirow{2}{*}{\parbox{2cm}{\centering \textbf{Text-trajectory alignment}\textuparrow}}
& \multirow{2}{*}{\parbox{1.5cm}{\centering \textbf{Reference fidelity}\textuparrow}}
& \multirow{2}{*}{\parbox{1.5cm}{\centering \textbf{Overall video quality}\textuparrow}} \\
& & & \\ 
\hline
Wan2.2 I2V & 1.67\% & \underline{11.00\%} & \underline{4.33\%} & - & \underline{19.67\%} \\
ATI & \underline{19.00\%} & 9.67\% & 3.33\% & - & 9.67\% \\
Frame In-N-Out & 4.00\% & 5.67\% & 3.33\% & \underline{7.33\%} & 1.33\% \\
\hline
\methodname (Ours) & \textbf{75.33\%} & \textbf{73.67\%} & \textbf{89.00\%} & \textbf{92.67\%} & \textbf{69.33\%} \\
\bottomrule
\end{tabular}
\end{table*}

%% file: sec/ref.tex
{
\small
\bibliographystyle{ieeenat_fullname}
\bibliography{main.bbl}
}

%% file: main.bbl
\begin{thebibliography}{48}
\providecommand{\natexlab}[1]{#1}
\providecommand{\url}[1]{\texttt{#1}}
\expandafter\ifx\csname urlstyle\endcsname\relax
  \providecommand{\doi}[1]{doi: #1}\else
  \providecommand{\doi}{doi: \begingroup \urlstyle{rm}\Url}\fi

\bibitem[Agarwal et~al.(2025)Agarwal, Ali, Bala, Balaji, Barker, Cai, Chattopadhyay, Chen, Cui, Ding, et~al.]{agarwal2025cosmos}
Niket Agarwal, Arslan Ali, Maciej Bala, Yogesh Balaji, Erik Barker, Tiffany Cai, Prithvijit Chattopadhyay, Yongxin Chen, Yin Cui, Yifan Ding, et~al.
\newblock Cosmos world foundation model platform for physical ai.
\newblock \emph{arXiv preprint arXiv:2501.03575}, 2025.

\bibitem[Bai et~al.(2025)Bai, Chen, Liu, Wang, Ge, Song, Dang, Wang, Wang, Tang, et~al.]{qwenvl}
Shuai Bai, Keqin Chen, Xuejing Liu, Jialin Wang, Wenbin Ge, Sibo Song, Kai Dang, Peng Wang, Shijie Wang, Jun Tang, et~al.
\newblock Qwen2. 5-vl technical report.
\newblock \emph{arXiv preprint arXiv:2502.13923}, 2025.

\bibitem[Ball et~al.(2025)Ball, Bauer, Belletti, Brownfield, Ephrat, Fruchter, Gupta, Holsheimer, Holynski, Hron, Kaplanis, Limont, McGill, Oliveira, Parker-Holder, Perbet, Scully, Shar, Spencer, Tov, Villegas, Wang, Yung, Baetu, Berbel, Bridson, Bruce, Buttimore, Chakera, Chandra, Collins, Cullum, Damoc, Dasagi, Gazeau, Gbadamosi, Han, Hirst, Kachra, Kerley, Kjems, Knoepfel, Koriakin, Lo, Lu, Mehring, Moufarek, Nandwani, Oliveira, Pardo, Park, Pierson, Poole, Ran, Salimans, Sanchez, Saprykin, Shen, Sidhwani, Smith, Stanton, Tomlinson, Vijaykumar, Wang, Wingfield, Wong, Xu, Yew, Young, Zubov, Eck, Erhan, Kavukcuoglu, Hassabis, Gharamani, Hadsell, van~den Oord, Mosseri, Bolton, Singh, and Rockt{\"a}schel]{genie3}
Philip~J. Ball, Jakob Bauer, Frank Belletti, Bethanie Brownfield, Ariel Ephrat, Shlomi Fruchter, Agrim Gupta, Kristian Holsheimer, Aleksander Holynski, Jiri Hron, Christos Kaplanis, Marjorie Limont, Matt McGill, Yanko Oliveira, Jack Parker-Holder, Frank Perbet, Guy Scully, Jeremy Shar, Stephen Spencer, Omer Tov, Ruben Villegas, Emma Wang, Jessica Yung, Cip Baetu, Jordi Berbel, David Bridson, Jake Bruce, Gavin Buttimore, Sarah Chakera, Bilva Chandra, Paul Collins, Alex Cullum, Bogdan Damoc, Vibha Dasagi, Maxime Gazeau, Charles Gbadamosi, Woohyun Han, Ed Hirst, Ashyana Kachra, Lucie Kerley, Kristian Kjems, Eva Knoepfel, Vika Koriakin, Jessica Lo, Cong Lu, Zeb Mehring, Alex Moufarek, Henna Nandwani, Valeria Oliveira, Fabio Pardo, Jane Park, Andrew Pierson, Ben Poole, Helen Ran, Tim Salimans, Manuel Sanchez, Igor Saprykin, Amy Shen, Sailesh Sidhwani, Duncan Smith, Joe Stanton, Hamish Tomlinson, Dimple Vijaykumar, Luyu Wang, Piers Wingfield, Nat Wong, Keyang Xu, Christopher Yew, Nick Young, Vadim Zubov, Douglas
  Eck, Dumitru Erhan, Koray Kavukcuoglu, Demis Hassabis, Zoubin Gharamani, Raia Hadsell, A{\"a}ron van~den Oord, Inbar Mosseri, Adrian Bolton, Satinder Singh, and Tim Rockt{\"a}schel.
\newblock Genie 3: A new frontier for world models.
\newblock 2025.

\bibitem[Blattmann et~al.(2023)Blattmann, Dockhorn, Kulal, Mendelevitch, Kilian, Lorenz, Levi, English, Voleti, Letts, et~al.]{blattmann2023stable}
Andreas Blattmann, Tim Dockhorn, Sumith Kulal, Daniel Mendelevitch, Maciej Kilian, Dominik Lorenz, Yam Levi, Zion English, Vikram Voleti, Adam Letts, et~al.
\newblock Stable video diffusion: Scaling latent video diffusion models to large datasets.
\newblock \emph{arXiv preprint arXiv:2311.15127}, 2023.

\bibitem[Bochkovskiy et~al.(2020)Bochkovskiy, Wang, and Liao]{yolo}
Alexey Bochkovskiy, Chien-Yao Wang, and Hong-Yuan~Mark Liao.
\newblock Yolov4: Optimal speed and accuracy of object detection.
\newblock \emph{arXiv preprint arXiv:2004.10934}, 2020.

\bibitem[Chu et~al.(2025)Chu, He, Chen, Zhang, Xu, Xia, Wang, Yi, Liu, Zhao, et~al.]{chuwan}
Ruihang Chu, Yefei He, Zhekai Chen, Shiwei Zhang, Xiaogang Xu, Bin Xia, Dingdong Wang, Hongwei Yi, Xihui Liu, Hengshuang Zhao, et~al.
\newblock Wan-move: Motion-controllable video generation via latent trajectory guidance.
\newblock In \emph{Adv. Neural Inform. Process. Syst.}, 2025.

\bibitem[Esser et~al.(2024)Esser, Kulal, Blattmann, Entezari, M{\"{u}}ller, Saini, Levi, Lorenz, Sauer, Boesel, Podell, Dockhorn, English, and Rombach]{flow2}
Patrick Esser, Sumith Kulal, Andreas Blattmann, Rahim Entezari, Jonas M{\"{u}}ller, Harry Saini, Yam Levi, Dominik Lorenz, Axel Sauer, Frederic Boesel, Dustin Podell, Tim Dockhorn, Zion English, and Robin Rombach.
\newblock Scaling rectified flow transformers for high-resolution image synthesis.
\newblock In \emph{Int. Conf. Mach. Learn.}, 2024.

\bibitem[Geng et~al.(2025)Geng, Herrmann, Hur, Cole, Zhang, Pfaff, Lopez-Guevara, Aytar, Rubinstein, Sun, et~al.]{geng2025motion}
Daniel Geng, Charles Herrmann, Junhwa Hur, Forrester Cole, Serena Zhang, Tobias Pfaff, Tatiana Lopez-Guevara, Yusuf Aytar, Michael Rubinstein, Chen Sun, et~al.
\newblock Motion prompting: Controlling video generation with motion trajectories.
\newblock In \emph{IEEE Conf. Comput. Vis. Pattern Recog.}, 2025.

\bibitem[Gillman et~al.(2025)Gillman, Herrmann, Freeman, Aggarwal, Luo, Sun, and Sun]{gillman2025force}
Nate Gillman, Charles Herrmann, Michael Freeman, Daksh Aggarwal, Evan Luo, Deqing Sun, and Chen Sun.
\newblock Force prompting: Video generation models can learn and generalize physics-based control signals.
\newblock \emph{arXiv preprint arXiv:2505.19386}, 2025.

\bibitem[Hafner et~al.(2023)Hafner, Pasukonis, Ba, and Lillicrap]{hafner2023mastering}
Danijar Hafner, Jurgis Pasukonis, Jimmy Ba, and Timothy Lillicrap.
\newblock Mastering diverse domains through world models.
\newblock \emph{arXiv preprint arXiv:2301.04104}, 2023.

\bibitem[He et~al.(2025{\natexlab{a}})He, Yang, Lin, Xu, Wei, Gui, Zhao, Wetzstein, Jiang, and Li]{he2025cameractrl}
Hao He, Ceyuan Yang, Shanchuan Lin, Yinghao Xu, Meng Wei, Liangke Gui, Qi Zhao, Gordon Wetzstein, Lu Jiang, and Hongsheng Li.
\newblock Cameractrl ii: Dynamic scene exploration via camera-controlled video diffusion models.
\newblock \emph{arXiv preprint arXiv:2503.10592}, 2025{\natexlab{a}}.

\bibitem[He et~al.(2025{\natexlab{b}})He, Peng, Liu, Wang, Zhang, Cui, Kang, Jiang, An, Ren, Xu, Guo, Gong, Wu, Li, Song, Liu, Li, and Zhou]{matrixgame2}
Xianglong He, Chunli Peng, Zexiang Liu, Boyang Wang, Yifan Zhang, Qi Cui, Fei Kang, Biao Jiang, Mengyin An, Yangyang Ren, Baixin Xu, Hao-Xiang Guo, Kaixiong Gong, Cyrus Wu, Wei Li, Xuchen Song, Yang Liu, Eric Li, and Yahui Zhou.
\newblock Matrix-game 2.0: An open-source, real-time, and streaming interactive world model.
\newblock \emph{arXiv preprint arXiv:2508.13009}, 2025{\natexlab{b}}.

\bibitem[Huang et~al.(2025)Huang, Wu, Zhou, Miao, and Long]{huang2025vid2world}
Siqiao Huang, Jialong Wu, Qixing Zhou, Shangchen Miao, and Mingsheng Long.
\newblock Vid2world: Crafting video diffusion models to interactive world models.
\newblock \emph{arXiv preprint arXiv:2505.14357}, 2025.

\bibitem[Huang et~al.(2024)Huang, He, Yu, Zhang, Si, Jiang, Zhang, Wu, Jin, Chanpaisit, Wang, Chen, Wang, Lin, Qiao, and Liu]{vbench}
Ziqi Huang, Yinan He, Jiashuo Yu, Fan Zhang, Chenyang Si, Yuming Jiang, Yuanhan Zhang, Tianxing Wu, Qingyang Jin, Nattapol Chanpaisit, Yaohui Wang, Xinyuan Chen, Limin Wang, Dahua Lin, Yu Qiao, and Ziwei Liu.
\newblock Vbench: Comprehensive benchmark suite for video generative models.
\newblock In \emph{IEEE Conf. Comput. Vis. Pattern Recog.}, pages 21807--21818. {IEEE}, 2024.

\bibitem[HunyuanWorld(2025)]{hunyuanworld}
Team HunyuanWorld.
\newblock Hunyuanworld 1.0: Generating immersive, explorable, and interactive 3d worlds from words or pixels.
\newblock \emph{arXiv preprint}, 2025.

\bibitem[Kang et~al.(2024)Kang, Yue, Lu, Lin, Zhao, Wang, Huang, and Feng]{kang2024far}
Bingyi Kang, Yang Yue, Rui Lu, Zhijie Lin, Yang Zhao, Kaixin Wang, Gao Huang, and Jiashi Feng.
\newblock How far is video generation from world model: A physical law perspective.
\newblock \emph{arXiv preprint arXiv:2411.02385}, 2024.

\bibitem[Karaev et~al.(2025)Karaev, Makarov, Wang, Neverova, Vedaldi, and Rupprecht]{cotracker3}
Nikita Karaev, Yuri Makarov, Jianyuan Wang, Natalia Neverova, Andrea Vedaldi, and Christian Rupprecht.
\newblock Cotracker3: Simpler and better point tracking by pseudo-labelling real videos.
\newblock In \emph{Int. Conf. Comput. Vis.}, 2025.

\bibitem[Kirillov et~al.(2023)Kirillov, Mintun, Ravi, Mao, Rolland, Gustafson, Xiao, Whitehead, Berg, Lo, Doll{\'{a}}r, and Girshick]{sam}
Alexander Kirillov, Eric Mintun, Nikhila Ravi, Hanzi Mao, Chlo{\'{e}} Rolland, Laura Gustafson, Tete Xiao, Spencer Whitehead, Alexander~C. Berg, Wan{-}Yen Lo, Piotr Doll{\'{a}}r, and Ross~B. Girshick.
\newblock Segment anything.
\newblock In \emph{Int. Conf. Comput. Vis.}, 2023.

\bibitem[Lei et~al.(2025)Lei, Wang, Zhang, Wang, Li, and Xu]{lei2025animateanything}
Guojun Lei, Chi Wang, Rong Zhang, Yikai Wang, Hong Li, and Weiwei Xu.
\newblock Animateanything: Consistent and controllable animation for video generation.
\newblock In \emph{IEEE Conf. Comput. Vis. Pattern Recog.}, 2025.

\bibitem[Lipman et~al.(2023)Lipman, Chen, Ben{-}Hamu, Nickel, and Le]{flow}
Yaron Lipman, Ricky T.~Q. Chen, Heli Ben{-}Hamu, Maximilian Nickel, and Matthew Le.
\newblock Flow matching for generative modeling.
\newblock In \emph{Int. Conf. Learn. Represent.}, 2023.

\bibitem[Liu et~al.(2024)Liu, Ren, Gupta, and Wang]{liu2024physgen}
Shaowei Liu, Zhongzheng Ren, Saurabh Gupta, and Shenlong Wang.
\newblock Physgen: Rigid-body physics-grounded image-to-video generation.
\newblock In \emph{Eur. Conf. Comput. Vis.}, 2024.

\bibitem[Liu et~al.(2025)Liu, Min, Wang, Wu, Wang, Yuan, Luo, and Guo]{worldmirror}
Yifan Liu, Zhiyuan Min, Zhenwei Wang, Junta Wu, Tengfei Wang, Yixuan Yuan, Yawei Luo, and Chunchao Guo.
\newblock Worldmirror: Universal 3d world reconstruction with any-prior prompting.
\newblock \emph{arXiv preprint arXiv:2510.10726}, 2025.

\bibitem[Ma et~al.(2023)Ma, Lewis, and Kleijn]{trailblazer}
Wan-Duo~Kurt Ma, J.~P. Lewis, and W.~Bastiaan Kleijn.
\newblock Trailblazer: Trajectory control for diffusion-based video generation.
\newblock \emph{arXiv preprint arXiv:2401.00896}, 2023.

\bibitem[Mao et~al.(2025)Mao, Jiang, Wang, Zhang, Chen, Chi, Wang, and Luo]{mao2025osv}
Xiaofeng Mao, Zhengkai Jiang, Fu-Yun Wang, Jiangning Zhang, Hao Chen, Mingmin Chi, Yabiao Wang, and Wenhan Luo.
\newblock Osv: One step is enough for high-quality image to video generation.
\newblock In \emph{IEEE Conf. Comput. Vis. Pattern Recog.}, 2025.

\bibitem[Namekata et~al.(2025)Namekata, Bahmani, Wu, Kant, Gilitschenski, and Lindell]{namekata2024sg}
Koichi Namekata, Sherwin Bahmani, Ziyi Wu, Yash Kant, Igor Gilitschenski, and David~B Lindell.
\newblock Sg-i2v: Self-guided trajectory control in image-to-video generation.
\newblock In \emph{Int. Conf. Learn. Represent.}, 2025.

\bibitem[Ouyang et~al.(2024)Ouyang, Wang, Xiao, Bai, Zhang, Zheng, Zhou, Chen, and Shen]{ouyang2024codef}
Hao Ouyang, Qiuyu Wang, Yuxi Xiao, Qingyan Bai, Juntao Zhang, Kecheng Zheng, Xiaowei Zhou, Qifeng Chen, and Yujun Shen.
\newblock Codef: Content deformation fields for temporally consistent video processing.
\newblock In \emph{IEEE Conf. Comput. Vis. Pattern Recog.}, 2024.

\bibitem[Po et~al.(2025)Po, Nitzan, Zhang, Chen, Dao, Shechtman, Wetzstein, and Huang]{po2025long}
Ryan Po, Yotam Nitzan, Richard Zhang, Berlin Chen, Tri Dao, Eli Shechtman, Gordon Wetzstein, and Xun Huang.
\newblock Long-context state-space video world models.
\newblock In \emph{Int. Conf. Comput. Vis.}, 2025.

\bibitem[Ren et~al.(2025)Ren, Wei, Guo, Zhao, Kang, Feng, and Jin]{ren2025videoworld}
Zhongwei Ren, Yunchao Wei, Xun Guo, Yao Zhao, Bingyi Kang, Jiashi Feng, and Xiaojie Jin.
\newblock Videoworld: Exploring knowledge learning from unlabeled videos.
\newblock In \emph{IEEE Conf. Comput. Vis. Pattern Recog.}, 2025.

\bibitem[Shen et~al.(2025)Shen, Jiang, Zhu, Ge, Cao, and Zheng]{shen2025identity}
Liao Shen, Wentao Jiang, Yiran Zhu, Tiezheng Ge, Zhiguo Cao, and Bo Zheng.
\newblock Identity-preserving image-to-video generation via reward-guided optimization.
\newblock \emph{arXiv preprint arXiv:2510.14255}, 2025.

\bibitem[Shi et~al.(2024)Shi, Huang, Wang, Bian, Li, Zhang, Zhang, Cheung, See, Qin, et~al.]{shi2024motion}
Xiaoyu Shi, Zhaoyang Huang, Fu-Yun Wang, Weikang Bian, Dasong Li, Yi Zhang, Manyuan Zhang, Ka~Chun Cheung, Simon See, Hongwei Qin, et~al.
\newblock Motion-i2v: Consistent and controllable image-to-video generation with explicit motion modeling.
\newblock In \emph{ACM SIGGRAPH Conference}, 2024.

\bibitem[Soucek and Lokoc(2024)]{transnetV2}
Tom{\'{a}}s Soucek and Jakub Lokoc.
\newblock Transnet {V2:} an effective deep network architecture for fast shot transition detection.
\newblock In \emph{ACM Int. Conf. Multimedia}, 2024.

\bibitem[Wan et~al.(2025)Wan, Wang, Ai, Wen, Mao, Xie, Chen, Yu, Zhao, Yang, et~al.]{Wan}
Team Wan, Ang Wang, Baole Ai, Bin Wen, Chaojie Mao, Chen-Wei Xie, Di Chen, Feiwu Yu, Haiming Zhao, Jianxiao Yang, et~al.
\newblock Wan: Open and advanced large-scale video generative models.
\newblock \emph{arXiv preprint arXiv:2503.20314}, 2025.

\bibitem[Wang et~al.(2025{\natexlab{a}})Wang, Huang, Fang, Yang, and Ma]{ATI}
Angtian Wang, Haibin Huang, Jacob~Zhiyuan Fang, Yiding Yang, and Chongyang Ma.
\newblock Ati: Any trajectory instruction for controllable video generation.
\newblock \emph{arXiv preprint arXiv:2505.22944}, 2025{\natexlab{a}}.

\bibitem[Wang et~al.(2025{\natexlab{b}})Wang, Chen, Gadelha, and Cheng]{frame-in-out}
Boyang Wang, Xuweiyi Chen, Matheus Gadelha, and Zezhou Cheng.
\newblock Frame in-n-out: Unbounded controllable image-to-video generation.
\newblock \emph{arXiv preprint arXiv:2505.21491}, 2025{\natexlab{b}}.

\bibitem[Wang et~al.(2025{\natexlab{c}})Wang, Ouyang, Wang, Wang, Cheng, Chen, Shen, and Wang]{Levitor}
Hanlin Wang, Hao Ouyang, Qiuyu Wang, Wen Wang, Ka~Leong Cheng, Qifeng Chen, Yujun Shen, and Limin Wang.
\newblock Levitor: 3d trajectory oriented image-to-video synthesis.
\newblock In \emph{IEEE Conf. Comput. Vis. Pattern Recog.}, 2025{\natexlab{c}}.

\bibitem[Wang et~al.(2023)Wang, He, Li, Li, Yu, Ma, Chen, Wang, Luo, Liu, Wang, Wang, and Qiao]{viclip}
Yi Wang, Yinan He, Yizhuo Li, Kunchang Li, Jiashuo Yu, Xin Ma, Xinyuan Chen, Yaohui Wang, Ping Luo, Ziwei Liu, Yali Wang, Limin Wang, and Yu Qiao.
\newblock Internvid: A large-scale video-text dataset for multimodal understanding and generation.
\newblock \emph{arXiv preprint arXiv:2307.06942}, 2023.

\bibitem[Wang et~al.(2024)Wang, Yuan, Wang, Li, Chen, Xia, Luo, and Shan]{motionctrl}
Zhouxia Wang, Ziyang Yuan, Xintao Wang, Yaowei Li, Tianshui Chen, Menghan Xia, Ping Luo, and Ying Shan.
\newblock Motionctrl: {A} unified and flexible motion controller for video generation.
\newblock In \emph{ACM SIGGRAPH Conference}, 2024.

\bibitem[Wiedemer et~al.(2025)Wiedemer, Li, Vicol, Gu, Matarese, Swersky, Kim, Jaini, and Geirhos]{wiedemer2025video}
Thadd{\"a}us Wiedemer, Yuxuan Li, Paul Vicol, Shixiang~Shane Gu, Nick Matarese, Kevin Swersky, Been Kim, Priyank Jaini, and Robert Geirhos.
\newblock Video models are zero-shot learners and reasoners.
\newblock \emph{arXiv preprint arXiv:2509.20328}, 2025.

\bibitem[Wu et~al.(2025)Wu, Yang, Po, Xu, Liu, Lin, and Wetzstein]{wu2025video}
Tong Wu, Shuai Yang, Ryan Po, Yinghao Xu, Ziwei Liu, Dahua Lin, and Gordon Wetzstein.
\newblock Video world models with long-term spatial memory.
\newblock \emph{arXiv preprint arXiv:2506.05284}, 2025.

\bibitem[Wu et~al.(2024)Wu, Li, Gu, Zhao, He, Zhang, Shou, Li, Gao, and Zhang]{DragAnything}
Weijia Wu, Zhuang Li, Yuchao Gu, Rui Zhao, Yefei He, David~Junhao Zhang, Mike~Zheng Shou, Yan Li, Tingting Gao, and Di Zhang.
\newblock Draganything: Motion control for anything using entity representation.
\newblock In \emph{Eur. Conf. Comput. Vis.}, 2024.

\bibitem[Yariv et~al.(2025)Yariv, Kirstain, Zohar, Sheynin, Taigman, Adi, Benaim, and Polyak]{yariv2025through}
Guy Yariv, Yuval Kirstain, Amit Zohar, Shelly Sheynin, Yaniv Taigman, Yossi Adi, Sagie Benaim, and Adam Polyak.
\newblock Through-the-mask: Mask-based motion trajectories for image-to-video generation.
\newblock In \emph{IEEE Conf. Comput. Vis. Pattern Recog.}, 2025.

\bibitem[Yin et~al.(2023)Yin, Wu, Liang, Shi, Li, Ming, and Duan]{DragNUWA}
Shengming Yin, Chenfei Wu, Jian Liang, Jie Shi, Houqiang Li, Gong Ming, and Nan Duan.
\newblock Dragnuwa: Fine-grained control in video generation by integrating text, image, and trajectory.
\newblock \emph{arXiv preprint arXiv:2308.08089}, 2023.

\bibitem[Zhang et~al.(2025{\natexlab{a}})Zhang, Wang, Fang, Jiang, Yang, Liu, Yang, Chen, Wen, Yuille, et~al.]{tgt}
Guofeng Zhang, Angtian Wang, Jacob~Zhiyuan Fang, Liming Jiang, Haotian Yang, Bo Liu, Yiding Yang, Guang Chen, Longyin Wen, Alan Yuille, et~al.
\newblock Tgt: Text-grounded trajectories for locally controlled video generation.
\newblock \emph{arXiv preprint arXiv:2510.15104}, 2025{\natexlab{a}}.

\bibitem[Zhang and Agrawala(2025)]{zhang2025packing}
Lvmin Zhang and Maneesh Agrawala.
\newblock Packing input frame context in next-frame prediction models for video generation.
\newblock \emph{arXiv preprint arXiv:2504.12626}, 2025.

\bibitem[Zhang et~al.(2023)Zhang, Wang, Zhang, Zhao, Yuan, Qin, Wang, Zhao, and Zhou]{zhang2023i2vgen}
Shiwei Zhang, Jiayu Wang, Yingya Zhang, Kang Zhao, Hangjie Yuan, Zhiwu Qin, Xiang Wang, Deli Zhao, and Jingren Zhou.
\newblock I2vgen-xl: High-quality image-to-video synthesis via cascaded diffusion models.
\newblock \emph{arXiv preprint arXiv:2311.04145}, 2023.

\bibitem[Zhang et~al.(2025{\natexlab{b}})Zhang, Peng, Wang, Wang, Zhu, Kang, Jiang, Gao, Li, Liu, and Zhou]{matrixgame}
Yifan Zhang, Chunli Peng, Boyang Wang, Puyi Wang, Qingcheng Zhu, Fei Kang, Biao Jiang, Zedong Gao, Eric Li, Yang Liu, and Yahui Zhou.
\newblock Matrix-game: Interactive world foundation model.
\newblock \emph{arXiv preprint arXiv:2506.18701}, 2025{\natexlab{b}}.

\bibitem[Zhang et~al.(2025{\natexlab{c}})Zhang, Liao, Li, Dai, Qiu, Zhu, Qin, and Wang]{Tora}
Zhenghao Zhang, Junchao Liao, Menghao Li, Zuozhuo Dai, Bingxue Qiu, Siyu Zhu, Long Qin, and Weizhi Wang.
\newblock Tora: Trajectory-oriented diffusion transformer for video generation.
\newblock In \emph{IEEE Conf. Comput. Vis. Pattern Recog.}, 2025{\natexlab{c}}.

\bibitem[Zhang et~al.(2025{\natexlab{d}})Zhang, Liao, Meng, Qin, and Wang]{to2}
Zhenghao Zhang, Junchao Liao, Xiangyu Meng, Long Qin, and Weizhi Wang.
\newblock Tora2: Motion and appearance customized diffusion transformer for multi-entity video generation.
\newblock \emph{arXiv preprint arXiv:2507.05963}, 2025{\natexlab{d}}.

\end{thebibliography}
